# COMBINING SEMANTIC AND SYNTACTIC STRUCTURE FOR LANGUAGE MODELING

*Rens Bod*

Informatics Research Institute, University of Leeds, Leeds LS2 9JT &
Institute for Logic, Language and Computation, University of Amsterdam
rens@scs.leeds.ac.uk

## ABSTRACT

Structured language models for speech recognition have been shown to remedy the weaknesses of $n$-gram models. All current structured language models, however, are limited in that they do not take into account dependencies between non-headwords. We show that non-headword dependencies contribute significantly to improved word error rate, and that a data-oriented parsing model trained on semantically and syntactically annotated data can exploit these dependencies. This paper contains the first published experiments with a data-oriented parsing model trained by means of a maximum likelihood reestimation procedure.

## 1. INTRODUCTION

Structured language models for speech recognition have recently gained a considerable interest. They have been shown to outperform the 3-gram language model on various domains and they can be efficiently parsed in a left-to-right manner (Chelba & Jelinek 1998; Chelba 2000). Although it has been reported that higher order $n$-gram models perform as well as structured language models (Goodman 2000), it is widely recognized that $n$-gram models have intrinsic limitations in that they cannot capture lexical dependencies that are structurally rather than sequentially related.

Chelba & Jelinek (1998) have proposed a head-lexicalized stochastic grammar as the basis for a structured language model. Their grammar associates each nonterminal with its lexical head as in Collins (1999). While Chelba & Jelinek's language model outperforms a deleted interpolation 3-gram model, their head-lexicalized grammar is also limited in that it cannot capture dependencies between *non-*headwords, such as *more* and *than* in Wall Street Journal phrases like *more people than cargo*, or *more couples exchanging rings in 1988 than in the previous year*, where neither *more* nor *than* are headwords of these phrases (see Chelba 2000: 12-16).

A model which does not apply any a priori restrictions on the lexical dependencies is the Data-Oriented Parsing or DOP model (e.g. Bod 1993, 1998a/b; Sima'an 1999). The DOP model learns a stochastic tree-substitution grammar (STSG) from a treebank by extracting all subtrees (up to a certain depth) from the treebank and assigning probabilities to the subtrees that are proportional to their empirical treebank frequencies. Since subtrees can be lexicalized at their frontiers with one or more words, DOP takes into account both headword and non-headword dependencies. For example, the dependency between *more* and *than* is captured by a subtree where *more* and *than* are the only frontier words. In Bod (2000a) we have shown that non-headword dependencies contribute to higher parse accuracy on the Wall Street Journal corpus (Marcus et al. 1993), resulting in improved performance over previous systems such as Collins (1999) or Charniak (2000).

In the current paper, we propose the DOP model as a language model for speech recognition and test it on the OVIS spoken language corpus (10,000 syntactically and semantically annotated sentences with corresponding word-graphs -- see Bod 1998a/b). While DOP performs well as a parsing model (Bod 2000a), it achieves rather disappointing results as a language model: it performs worse than the 3-gram model on OVIS word-graphs. We therefore developed a new DOP model that reestimates the subtree probabilities by means of maximum likelihood training. This maximum likelihood DOP model does outperform the 3-gram model. We show that the elimination of subtrees with two or more non-headwords leads to a significant deterioration of the word error rate (WER). We also investigate the contribution of semantic information on the WER.

The rest of this paper is organized as follows. We first shortly describe the problem of language modeling in speech recognition. We then explain the DOP approach and show how it can be trained by means of a maximum likelihood reestimation procedure belonging to the class of expectation-maximization algorithms. Finally, we give an experimental evaluation of the various language models on the OVIS corpus.

## 2. LANGUAGE MODELS

In the statistical formulation of the speech recognition problem, the recognizer aims to find the word string

$$\hat{W} = \arg \max_W P(W) \, P(A \mid W)$$

where $A$ is the acoustic signal, $P(A \mid W)$ is the acoustic probability that when $W$ is spoken the signal $A$ results, and $P(W)$ is the a priori probability that the speaker will utter $W$. It is the task of a *language model* to estimate $P(W)$, while the *acoustic model* estimates $P(A \mid W)$. Most language models estimate the probability of $W = w_1, w_2, ..., w_n$ by the well-known $n$-gram model

$$P(W) = \prod_i P(w_i \mid w_{i-n+1}, w_{i-n+2}, ..., w_{i-1})$$

where in most cases $n = 3$. The shortcomings of the 3-gram model are well-known: it cannot capture lexical dependencies that are structurally rather than linearly related, such as between *dog* and *barked* in the sentence *The dog on the hill barked*. Morever, *dog* and *barked* can be arbitrarily widely separated, such as in *The dog near the house on the hill barked*, etc. These dependencies can only be captured by a structured language model, such as a head-lexicalized stochastic grammar which associates each nonterminal of a context-free rule with a lexical head (Chelba & Jelinek 1998). Since *dog* and *barked* are both phrasal headwords, a head-lexicalized grammar can capture the dependency between these two words. However, as said in the introduction, head-lexicalized grammars run into problems if there are dependencies between words that are *not* headwords of phrases, such as *more* and *than* in the Wall Street Journal construction *more people than cargo*, where the relevant dependency is between *more* and *than* rather than between *people* and *cargo*, but neither *more* nor *than* are taken as headwords by a head-lexicalized grammar (see Chelba 2000: 12-16). Moreover, non-headwords can be arbitrarily widely separated as in *more couples exchanging rings in 1988 than in the previous year*. Non-headword dependencies can only be captured by a model which does not apply a priori restrictions on the lexical dependencies, such as the Data-Oriented Parsing (DOP) model.

## 3. A DOP-BASED LANGUAGE MODEL

The DOP model presented in Bod (1993, 1998a) learns a stochastic tree-substitution grammar (STSG) from a treebank by taking all subtrees seen in that treebank to form a grammar. A substitution operation is used to combine subtrees into parse trees for new sentences (see Bod 2000b for some alternative DOP models). As an example, consider a very simple treebank consisting of only two trees:

(1)

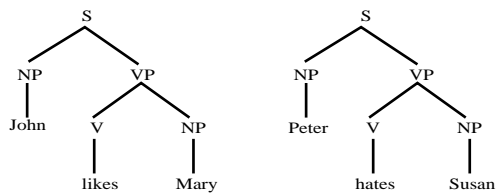

New sentences may be derived by combining subtrees from this treebank by means of a node-substitution operation indicated as ∘. Node-substitution identifies the leftmost nonterminal frontier node of one tree with the root node of a second tree. Under the convention that node-substitution is left-associative, a new sentence such as *Mary likes Susan* can be derived by combining subtrees from this treebank, as in (2):

(2)

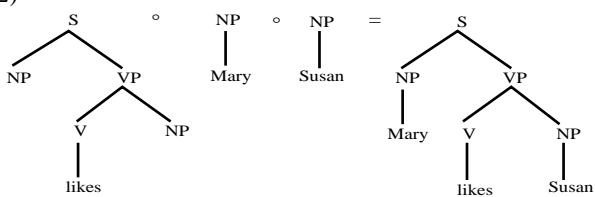

Other derivations may yield the same parse tree; for instance:

(3)

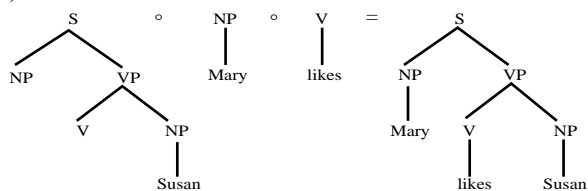

The DOP model in Bod (1993) computes the probability of a subtree $t$ as the probability of selecting $t$ among all corpus-subtrees that can be substituted on the same node as $t$. This probability is equal to the number of occurrences of $t$, $|t|$, divided by the total number of occurrences of all subtrees $t'$ with the same root label as $t$. Let $r(t)$ return the root label of $t$. Then we may write:

$$P(t) = \frac{|t|}{\sum_{t': r(t')=r(t)} |t'|}$$

The probability of a derivation $D = t_1 \circ ... \circ t_n$ is computed by the product of the probabilities of its subtrees $t_i$:

$$P(t_1 \circ ... \circ t_n) = \prod_i P(t_i)$$

As we have seen, there may be several distinct derivations that generate the same parse tree. The probability of a parse tree $T$ is thus the sum of the probabilities of all derivations $D$ that generate $T$:

$$P(T) = \sum_D P(T, D)$$

To use DOP as a language model for speech, we need to compute the probability of a word string $W$. Since a word string may have several different parse trees $T$, the probability of $W$ equals the sum of the probabilities of all parse trees $T$ that yield $W$ (which is also equal to the sum of the probabilities of all derivations $D$ that generate $W$):

$$P(W) = \sum_T P(W, T)$$

Note that the DOP model considers counts of subtrees of a wide range of sizes and lexicalizations in computing the probability of a tree: everything from counts of single-level rules to counts of entire trees This means that the model is sensitive to the frequency of large subtrees while taking into account the smoothing effects of counts of small subtrees.

Although in figures (2) and (3) the subtrees are lexicalized with only one word, there is no upper bound in DOP on the number of words in a subtree frontier (though for computational reasons the size of the subtrees is sometimes restricted). This means that DOP can capture dependencies between headwords as well as between non-headwords. For example, the previously mentioned dependency between the non-headwords *more* and *than* is captured by a subtree in which *more* and *than* are the only frontier words.

Johnson (1998) has pointed out that the way DOP assigns probabilities to subtrees (by the relative frequency estimator) does not maximize the likelihood of the corpus. Although this may not seem very important from a performance point of view, as DOP obtains state-of-the-art parsing accuracy on the Wall Street Journal corpus (Bod 2000a), we will test in this paper also a DOP model which reestimates the subtree probabilities by a maximum likelihood reestimation procedure belonging to the class of expectation-maximization (EM) algorithms (Dempster et al. 1977).[1] We will refer to this new, maximum likelihood DOP model as "ML-DOP". Unfortunately, reestimation with ML-DOP is computationally very expensive because of the large number of parameters. We have therefore only very recently been able to train an ML-DOP model on a non-trivial corpus. The exposition of the following reestimation technique for DOP is heavily based on a technical report by Magerman (1993).

---

[1] Bonnema et al. (1999) present an alternative probability model which estimates the probability of a subtree as the probability that it has been involved in the derivation of a corpus tree. It is not yet known how this estimator compares to the EM algorithm.

It is important to realize that there is an implicit assumption in DOP that all derivations of a parse tree contribute equally to the total probability of the parse tree. This is equivalent to saying that there is a hidden component to the model, and that DOP can be trained using an EM algorithm to determine the maximum likelihood estimate for the training data. The EM algorithm for this new ML-DOP model is related to the Inside-Outside algorithm for context-free grammars, but the reestimation formula is complicated by the presence of subtrees of depth greater than 1. To derive the correct reestimation formula, it is useful to consider the state space of all possible derivations of a tree.

The derivations of a parse tree $T$ can be viewed as a state trellis, where each state contains a partially constructed tree in the course of a leftmost derivation of $T$. $s_t$ denotes a state containing the tree $t$ which is a subtree of $T$. The state trellis is defined as such:

The initial state, $s_0$, is a tree with depth zero, consisting of simply a root node labeled with $S$.

The final state, $s_T$, is the given parse tree $T$.

A state $s_t$ is connected forward to all states $s_{t_f}$ such that $t_f = t \circ t'$, for some $t'$. Here the appropriate $t'$ is defined to be $t_f - t$.

A state $s_t$ is connected backward to all states $s_{t_b}$ such that $t = t_b \circ t'$, for some $t'$. Again, $t'$ is defined to be $t - t_b$.

The construction of the state lattice and assignment of transition probabilities according to the ML-DOP model is called the forward pass. The probability of a given state, $P(s)$, is referred to as $\alpha(s)$. The forward probability of a state $s_t$ is computed recursively

$$\alpha(s_t) = \sum_{s_{t_b}} \alpha(s_{t_b}) P(t - t_b).$$

The backward probability of a state, referred to as $\beta(s)$, is calculated according to the following recursive formula:

$$\beta(s_t) = \sum_{s_{t_f}} \beta(s_{t_f}) P(t_f - t)$$

where the backward probability of the goal state is set equal to the forward probability of the goal state, $\beta(s_T) = \alpha(s_T)$.

The update formula for the count of a subtree $t$ is (where $r(t)$ is the root label of $t$):

$$ct(t) = \sum_{s_{t_b}: \exists s_{t_f}, t_b \circ t = t_f} \frac{\beta(s_{t_f}) \alpha(s_{t_b}) P(t \mid r(t))}{\alpha(s_{goal})}$$

The updated probability distribution, $P'(t \mid r(t))$, is defined to be

$$P'(t \mid r(t)) = \frac{ct(t)}{ct(r(t))}$$

where $ct(r(t))$ is defined as

$$ct(r(t)) = \sum_{t': r(t')=r(t)} ct(t')$$

## 4. COMPARING ML-DOP TO THE 3-GRAM MODEL ON THE OVIS CORPUS

The OVIS corpus consists of 10,000 user utterances (and corresponding word-graphs) about Dutch public transport information that are syntactically and semantically annotated. Each syntactic label is compositionally enriched with a semantic formula as described in Bonnema et al. (1997) and Bod (1998b, 1999). The following figure gives an example annotation for the OVIS sentence *Ik wil niet vandaag maar morgen naar Almere* (literally: "I want not today but tomorrow to Almere"). As can be seen, the pre-lexical nodes contain the meanings of the underlying lexical items, while the higher nodes contain a formula scheme indicating how the meaning of the constituent is built up out of the meanings of its subconstituents. These schemes use the variable *d1* to indicate the meaning of the leftmost daughter constituent, *d2* to indicate the meaning of the second daughter node constituent, etc. Moreover, the "#" in the example refers to the denied information while the "!" refers to the corrected information. For more details on the OVIS corpus and its annotation convention, see Bonnema et al. (1997) or Bod (1998a, 1999).

(4)

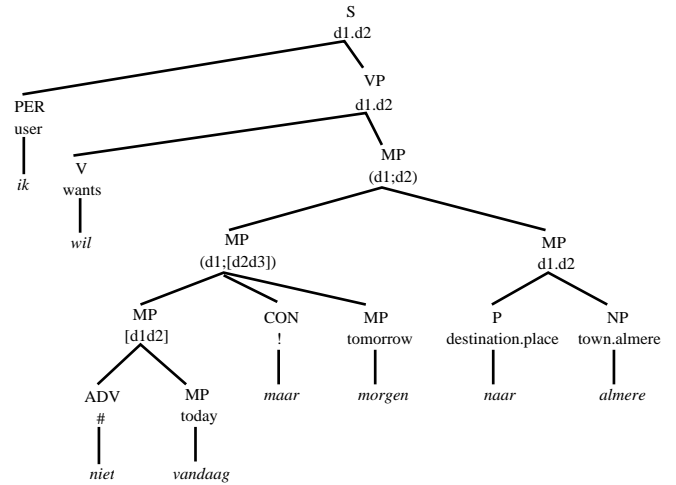

Thus the meaning of *Ik wil niet vandaag maar morgen naar Almere* is compositionally built up out of the meanings of its subconstituents. Substituting the meaning representations into the corresponding variables yields the semantics of the top-node S: `user.wants.(([# today];[! tomorrow]); destination.place.town.almere)`. An advantage of using compositional semantic annotations is that the treebank subtrees can directly be extracted and employed by the DOP model for computing syntactic/semantic structures for new sentences (Bod 1998a/b).

So far, the OVIS corpus has mainly been used by DOP to compute the most probable semantic interpretation from a word-graph by means of a Viterbi-style bottom-up parsing algorithm (see e.g. Bod 1998b). However, the DOP model may just as well be used to compute the most probable word string from each word-graph, by summing up the probabilities of all derivations that yield the same string. Since the computation of the most probable string by DOP is NP-hard (Sima'an 1996), we employ a Viterbi $n$ best search and estimate the most probable string by the 1,000 most probable derivations. If DOP does not find a derivation for a complete word-graph path, we compute the 1,000 most probable combinations of subderivations of partial paths. For further details on word-graph parsing with DOP we refer to Bod (1998b) or Sima'an (1999).

We split the OVIS corpus into 10 random divisions of 90%/10% training/test data, and used each training set to derive resp. the 3-gram model (using Katz smoothing -- see Katz 1987), the original DOP model, and the ML-DOP model. The original DOP model was obtained by extracting all subtrees up to depth 4 from the training set and assigning them probabilities proportional to their empirical relative frequencies. For the ML-DOP model, we also extracted all subtrees up to depth 4 from the training set which were then trained on the training set trees by maximum likelihood reestimation. The updated probabilities were iteratively reestimated until the decrease in cross-entropy became negligible. We never needed more than 30 iterations to obtain good convergence. The 1000 word-graphs for each test set were used as input, while the user utterances were kept apart. The word error rate (WER) for each model was averaged over all 10 test sets. The following table shows the results (where SimpleDOP refers to the original DOP model based on simple relative frequencies).

| Model | WER |
| --- | --- |
| 3-gram | 18.7% |
| SimpleDOP | 19.8% |
| ML-DOP | 17.0% |

WERs of the language models for OVIS word-graphs

The table shows that the 3-gram model outperforms the simple DOP model with 1.1% WER. This difference is statistically significant according to paired *t*-testing. However, ML-DOP outperforms the 3-gram model with 1.7% WER, which was also statistically significant. We next tested a version of ML-DOP where we eliminated all subtrees with two or more non-headwords. This led to a statistically significant deterioration of 0.9% WER. This result indicates that non-headword dependencies are important for predicting the uttered string, and should not be discarded as in head-lexicalized grammars (Chelba 2000; Collins 1999, 2000; Charniak 2000). Finally, we tested a version of ML-DOP where we eliminated the semantic annotations; this also led to a deterioration in WER of 1.3% which was statistically significant. Thus, compositional semantic annotations contribute to significantly better word error rate.

## 5. CONCLUSIONS

Previous structured language models did not take into account non-headword dependencies. We have shown that a maximum likelihood DOP model can effectively exploit non-headword dependencies, which significantly contribute to improved word error rate on the OVIS spoken language corpus. We have also seen that the maximum likelihood DOP model outperforms the 3-gram model, while the original DOP model performs worse than the 3-gram model. Finally, we found that semantic annotations contribute to significantly better word error rate. As future research, we want to test ML-DOP on Wall Street Journal utterances. This may be especially interesting as the DOP model has been shown to obtain state-of-the-art parsing performance on the WSJ (see Bod 2000a).

## REFERENCES

R. Bod, 1993. Using an Annotated Language Corpus as a Virtual Stochastic Grammar, *Proceedings AAAI'93*, Washington D.C.

R. Bod, 1998a. *Beyond Grammar: An Experience-Based Theory of Language,* CSLI Publications, distributed by Cambridge University Press.

R. Bod, 1998b. Spoken Dialogue Interpretation with the DOP Model. *Proceedings COLING-ACL'98*, Montreal, Canada.

R. Bod, 1999. Context-Sensitive Dialogue Processing with the DOP model, *Natural Language Engineering* 5(4), 309-323.

R. Bod, 2000a. Parsing with the Shortest Derivation, *Proceedings COLING'2000*, Saarbrücken, Germany.

R. Bod, 2000b. An Improved Parser for Data-Oriented Lexical-Functional Analysis. *Proceedings ACL'2000*, Hong Kong, China.

R. Bonnema, R. Bod and R. Scha, 1997. A DOP Model for Semantic Interpretation, *Proceedings ACL/EACL-97*, Madrid, Spain.

R. Bonnema, P. Buying and R. Scha, 1999. A New Probability Model for Data-Oriented Parsing. *Proceedings of the Amsterdam Colloquium 1999.* Amsterdam.

E. Charniak, 2000. A Maximum-Entropy-Inspired Parser. *Proceedings ANLP-NAACL'2000*, Seattle, Washington.

C. Chelba, 2000. *Exploiting Syntactic Structure for Natural Language Modeling*, PhD-thesis, Johns Hopkins University, Maryland.

C. Chelba and F. Jelinek, 1998. Exploiting Syntactic Structure for Natural Language Modeling, *Proceedings COLING-ACL'98*, Montreal, Canada.

M. Collins, 1999. *Head-Driven Statistical Models for Natural Language Parsing*, PhD-thesis, University of Pennsylvania, PA.

M. Collins, 2000. Discriminative Reranking for Natural Language Parsing, *Proceedings ICML'2000*, Stanford, Ca.

A. Dempster, N. Laird and D. Rubin, 1977. Maximum Likelihood from Incomplete Data via the EM Algorithm, *Journal of the Royal Statistical Society*, 39:1-38.

J. Goodman, 2000. *The State-of-the-Art in Language Modeling*, tutorial notes, ANLP-NAACL'2000, Seattle, Washington.

M. Johnson, 1998. The DOP Estimation Method is Biased and Inconsistent, squib.

S. Katz, 1987. Estimation of Probabilities from Sparse Data for the Language Model Component of a Speech Recognizer. *IEEE Transactions on Acoustics, Speech and Signal Processing*, 35:400-01.

D. Magerman, 1993. *Expectation-Maximization for Data-Oriented Parsing*, IBM Technical Report, Yorktown Heights, NY.

M. Marcus, B. Santorini and M. Marcinkiewicz, 1993. Building a Large Annotated Corpus of English: the Penn Treebank, *Computational Linguistics* 19(2), 313-330.

K. Sima'an, 1996. Computational Complexity of Probabilistic Disambiguation by means of Tree Grammars, *Proceedings COLING-96*, Copenhagen, Denmark.

K. Sima'an, 1999. *Learning Efficient Disambiguation*. PhD thesis, ILLC dissertation series number 1999-02. Utrecht/Amsterdam, The Netherlands.